%% file: main.tex
\documentclass[sigconf]{aamas}

\usepackage{balance} 
\usepackage{dsfont}
\usepackage{amsmath,amsfonts,amsthm}
\usepackage{algorithm}
\usepackage[noend]{algpseudocode} 
\usepackage{graphicx}
\usepackage{textcomp}
\usepackage{subcaption}
\usepackage[binary-units]{siunitx} 
\usepackage{cleveref} 
\usepackage{bbm} 
\usepackage{nicefrac}  
\usepackage{enumerate}  
\usepackage{booktabs} 
\usepackage{wrapfig}  

\hypersetup{bookmarksopen,bookmarksnumbered,
pdfpagemode=UseOutlines,
colorlinks=true,
linkcolor=red,
anchorcolor=red,
citecolor=red,
filecolor=red,
menucolor=red,
urlcolor=black
}

\setcopyright{ifaamas}
\acmConference[AAMAS '22]{Proc.\@ of the 21st International Conference
on Autonomous Agents and Multiagent Systems (AAMAS 2022)}{May 9--13, 2022}
{Online}{P.~Faliszewski, V.~Mascardi, C.~Pelachaud,
M.E.~Taylor (eds.)}
\copyrightyear{2022}
\acmYear{2022}
\acmDOI{}
\acmPrice{}
\acmISBN{}

\acmSubmissionID{103}

\title{Coordinated Multi-Agent Pathfinding for \\ Drones and Trucks over Road Networks}

\author{Shushman Choudhury}
\affiliation{
  \institution{Stanford University}
  \city{Stanford}
  \state{CA}
  \country{USA}}
\email{shushman@cs.stanford.edu}

\author{Kiril Solovey}
\affiliation{
  \institution{Technion - Israel Institute of Technology}
  \city{Haifa}
  \country{Israel}}
\email{kirilsol@stanford.edu}

\author{Mykel Kochenderfer}
\affiliation{
  \institution{Stanford University}
  \city{Stanford}
  \state{CA}
  \country{USA}}
\email{mykel@stanford.edu}

\author{Marco Pavone}
\affiliation{
  \institution{Stanford University}
  \city{Stanford}
  \state{CA}
  \country{USA}}
\email{pavone@stanford.edu}

\begin{abstract}
We address the problem of routing a team of drones and trucks over large-scale urban road networks. To conserve their limited flight energy, drones can use trucks as temporary modes of transit en route to their own destinations. Such coordination can yield significant savings in total vehicle distance traveled, i.e., truck travel distance and drone flight distance, compared to operating drones and trucks independently. But it comes at the potentially prohibitive computational cost of deciding which trucks and drones should coordinate and when and where it is most beneficial to do so. We tackle this fundamental trade-off by decoupling our overall intractable problem into tractable sub-problems that we solve stage-wise. The first stage solves only for trucks, by computing paths that make them more likely to be useful transit options for drones. The second stage solves only for drones, by routing them over a composite of the road network and the transit network defined by truck paths from the first stage. We design a comprehensive algorithmic framework that frames each stage as a multi-agent path-finding problem and implement two distinct methods for solving them. We evaluate our approach on extensive simulations with up to $100$ agents on the real-world Manhattan road network containing nearly $4500$ vertices and $10000$ edges. Our framework saves on more than $50\%$ of vehicle distance traveled compared to independently solving for trucks and drones, and computes solutions for all settings within $5$ minutes on commodity hardware.
\end{abstract}


\newcommand{\BibTeX}{\rm B\kern-.05em{\sc i\kern-.025em b}\kern-.08em\TeX}

\begin{document}

\pagestyle{fancy}
\fancyhead{}

\maketitle

\input{tex/macros}

\arxivfalse

\graphicspath{{fig/}}

\input{tex/introduction}

\input{tex/related}

\input{tex/problem}

\input{tex/approach}

\input{tex/experiments}

\input{tex/conclusion}

\begin{acks}
This  work  was  supported  in  part  by  the  National  Science  Foundation  award  no.\
1830554, the Toyota Research Institute, the Israeli Ministry of Science and Technology grants no.\ 3-16079 and 3-17385, and the United States-Israel Binational Science Foundation grants no.\ 2019703 and 2021643.
The authors thank Oren Salzman for insightful comments and suggestions.
\end{acks}

\balance

\bibliographystyle{ACM-Reference-Format}
\bibliography{tex/bibliography}


\end{document}

%% file: tex/macros.tex
\def\P{\mathcal{P}} \def\C{\mathcal{C}} \def\H{\mathcal{H}}
\def\F{\mathcal{F}} \def\U{\mathcal{U}} \def\L{\mathcal{L}}
\def\O{\mathcal{O}} \def\I{\mathcal{I}} \def\S{\mathcal{S}}
\def\G{\mathcal{G}} \def\Q{\mathcal{Q}} \def\I{\mathcal{I}}
\def\T{\mathcal{T}} \def\L{\mathcal{L}} \def\N{\mathcal{N}}
\def\V{\mathcal{V}} \def\B{\mathcal{B}} \def\D{\mathcal{D}}
\def\W{\mathcal{W}} \def\R{\mathcal{R}} \def\M{\mathcal{M}}
\def\X{\mathcal{X}} \def\A{\mathcal{A}} \def\Y{\mathcal{Y}}
\def\L{\mathcal{L}} \def\E{\mathcal{E}}

\def\dN{\mathbb{N}}

\def\tr{{\textup{tr}}}
\def\dr{{\textup{dr}}}

\def\Reals{\mathbb{R}}
\def\Naturals{\mathbb{N}}

\def\indicator{\mathds{1}}

\newtheorem{claim}{Claim}
\theoremstyle{definition}
\newtheorem{problem}{Problem}
\newtheorem{remark}{Remark}
\theoremstyle{plain}
\newtheorem{observation}{Observation}

\def\x{\bm{x}}

\newif\ifarxiv

\ifarxiv
\newcommand{\supp}[1]{Appendix #1}
\else
\newcommand{\supp}[1]{the appendix}
\fi

\newcommand{\numtrucks}{m}
\newcommand{\numdrones}{n}
\newcommand{\trucks}{[m]}
\newcommand{\drones}{[n]}

\newcommand{\ks}[2]{{\color{blue}#1}\marginpar{\color{blue}\raggedright\footnotesize [KS]:#2}}
\newcommand{\ksline}[2]{{\color{blue}#1}{\em \color{blue}[KS]: #2}}
\newcommand{\sh}[2]{{\color{red}#1}\marginpar{\color{red}\raggedright\footnotesize [SC]:#2}}
\newcommand{\new}[1]{{\color{blue}#1}}


%% file: tex/introduction.tex
\section{Introduction}
\label{sec:intro}

\begin{figure*}[t!]
    \centering
    \begin{subfigure}{0.45\textwidth}
        \centering
        \includegraphics[width=0.95\textwidth]{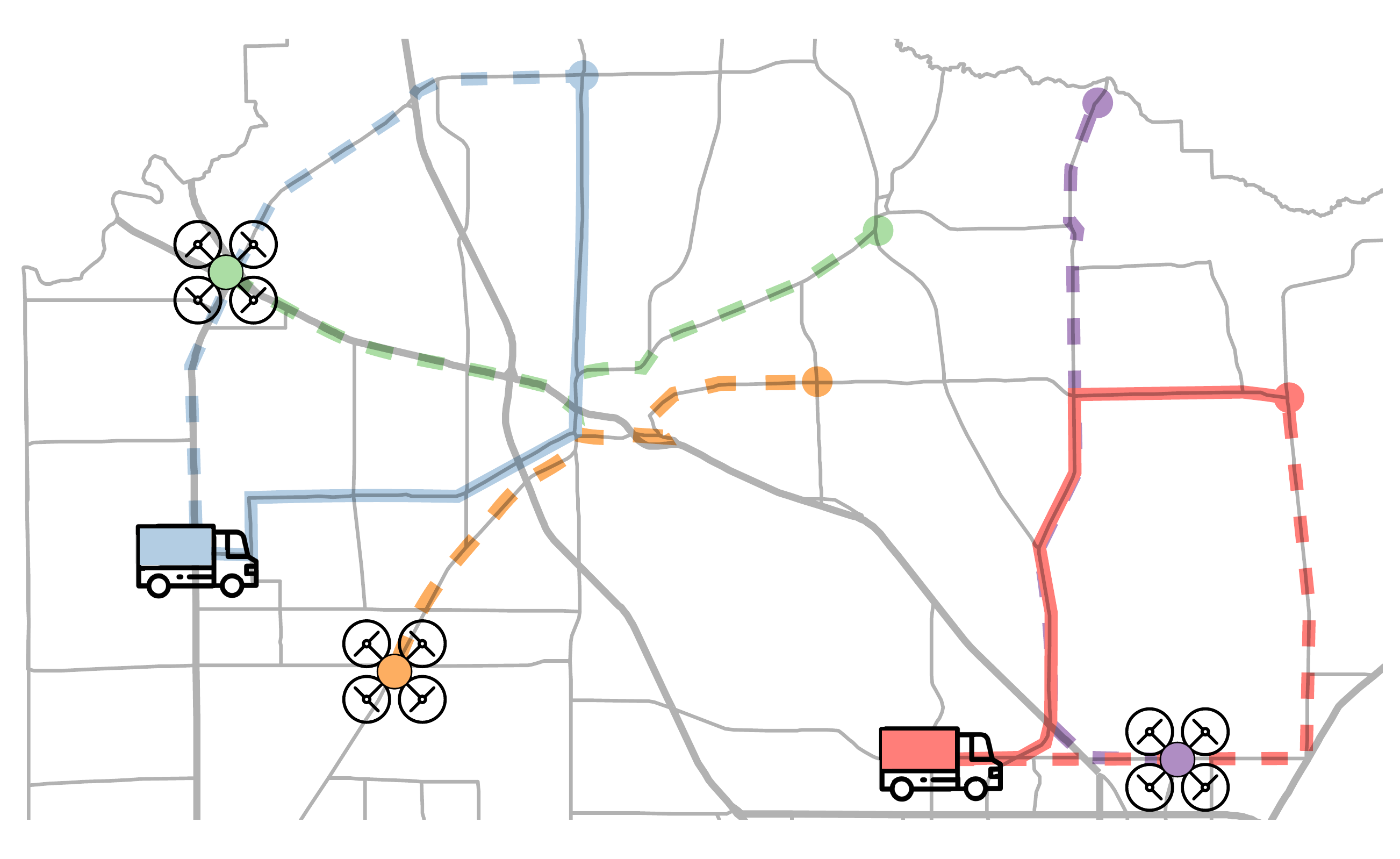}
    \end{subfigure}
    \begin{subfigure}{0.45\textwidth}
        \centering
        \includegraphics[width=0.95\textwidth]{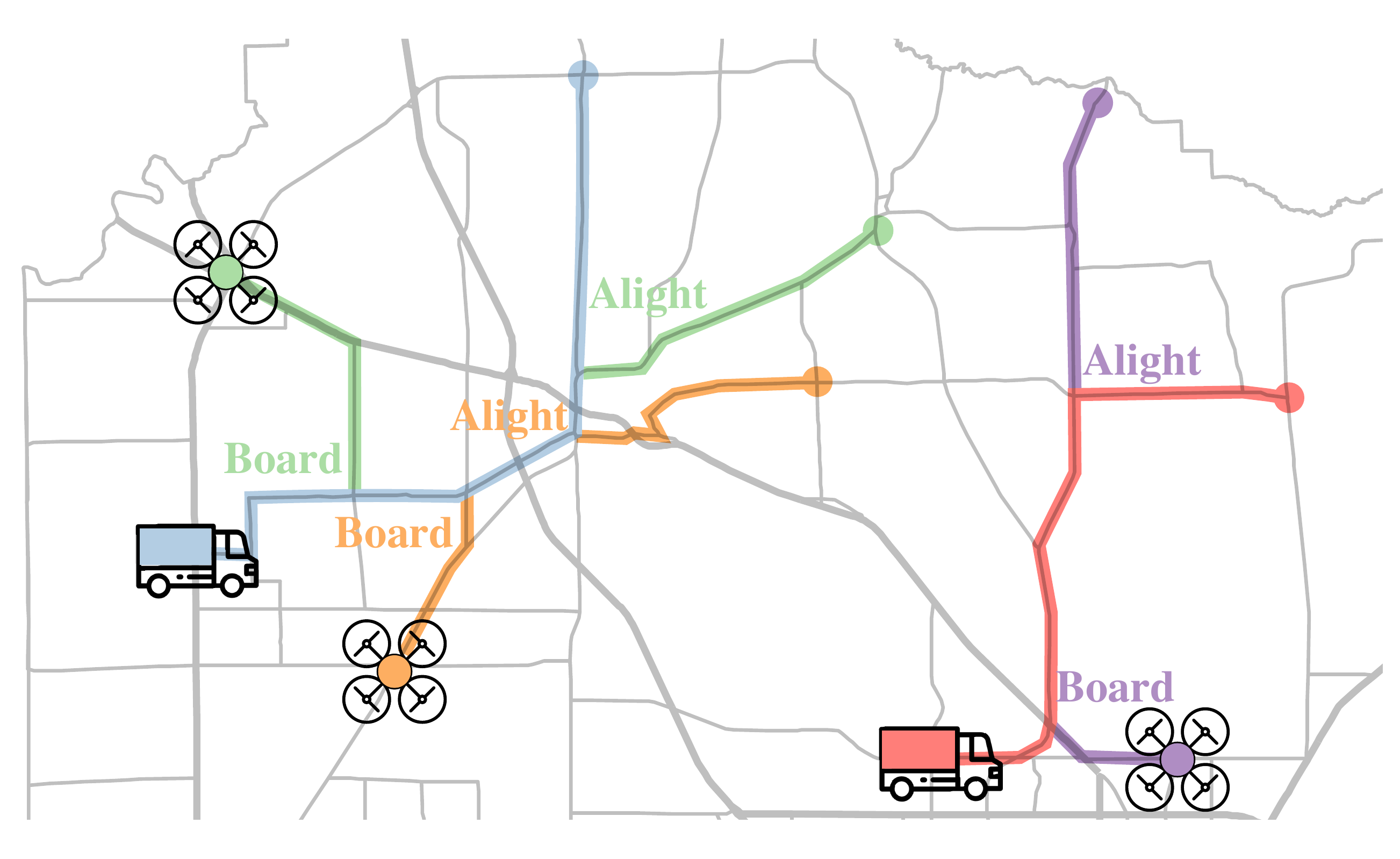}
    \end{subfigure}
    \caption{An illustration of our overall approach. (Left) Stage 1 computes truck paths solid (red and blue) in the vicinity of the shortest road paths for drones (dashed); these truck paths may deviate from the shortest road paths for the trucks (dashed red and blue). (Right) Stage 2 computes the shortest road-and-transit paths for drones, which can use trucks as transit. See~\Cref{sec:approach} for further details.} 
    \label{fig:approach}
\end{figure*}

Drones have great potential for transforming urban logistics services. By enabling quick, flexible, and efficient delivery, they can help address the rapidly growing logistics and e-commerce needs of dense urban populations~\cite{chung2020optimization}. They can also reduce our reliance on traditional ground delivery services that contribute to traffic congestion~\cite{HolguinETAL18}. However, operating package delivery services that rely solely on drones may be infeasible
due to their limited flight range and carrying capacity~\cite{SudburyETAL16}.
To overcome these limitations, we study the problem of operating delivery drones in tandem with ground vehicles by allowing drones to ride on ground vehicles to conserve energy and increase effective flight range~\cite{choudhury2019dynamic,choudhury2021efficient}.

In particular, we focus on routing a team of drones and trucks over a common road network, where drones can use trucks as transit in addition to flying. We frame our setting as a \emph{coordinated} extension of the Multi-Agent Path Finding (MAPF) problem~\cite{stern2019multi}, which requires us to compute start-goal routes for all agents while aiming to minimize the total path cost incurred by the drones and trucks. The path cost can be different for the two agent types and can encode the energy consumed or the operational expense. As in the classical MAPF formulation, our problem setting requires us to satisfy inter-agent constraints, which in our case bounds the maximum number of drones that can simultaneously use a truck. It also requires reasoning about the potential cost savings from coordinating trucks and drones to share trip segments. 

The feature of agents temporarily coordinating has not been explored by the MAPF community so far (to the best of our knowledge) and makes our problem much harder than the already difficult classical MAPF~\cite{yu2013structure, YuLaValle16}. 
Most state-of-the-art MAPF algorithms focus on avoiding inter-agent collisions rather than optimizing coordination~\cite{felner2017search,stern2019multi}. 
Our recent work developed a MAPF solver for routing drones over transit networks~\cite{choudhury2021efficient} but assumed the transit vehicles follow fixed and known routes and that drones were not required to fly over the road network. Another work considered MAPF problems with cooperation but only planned for predefined agent pairs to arrive at fixed meeting points~\cite{greshler2021cooperative}, rather than simultaneously traversing a route. The operations research community has looked at optimizing drones with trucks but their approaches only work for few agents on small abstract routing graphs~\cite{agatz2012optimization}.
\vspace{5pt}

\noindent\textbf{Contributions.} 
We develop an effective algorithmic approach for coordinated MAPF in the context of drones and ground trucks working in tandem. 
Our key idea is to \textit{decouple our overall intractable MAPF problem into two distinct MAPF sub-problems that we can solve in stages} (\Cref{fig:approach}). In Stage 1, we compute truck routes that are likely to be useful as transit options for drones. In Stage 2, we fix the truck routes, create a transit network based on them, and overlay this network on the road graph. We then compute drone routes over the composite road-and-transit network, where drones incur no  cost on the segments where they use transit (subject to capacity constraints). As a post-processing step, we re-route trucks not used as transit to their shortest road network paths.

We implement two variants of our approach that use different MAPF solvers; the first uses Enhanced Conflict-Based Search~\cite{barer2014suboptimal} and the second uses Prioritized Planning~\cite{silver2005cooperative} for both stages.
We evaluate our approach on a range of coordinated MAPF settings over the Manhattan road network, with nearly $4500$ vertices and $10000$ edges covering an area of nearly \SI{200}{\kilo\metre\squared}, and up to $20$ truck and $80$ drones. Our experiments show that coordinating drones and trucks can save more than $50\%$ of total vehicle distance traveled compared to no coordination. \textit{We refer to the sum of truck path distance and drone flight distance as vehicle distance traveled}, since drones incur no energy cost when riding on a truck. For brevity and convenience, we will use this phrase hereafter, with some abuse of terminology. Our approach plans paths for all settings within at most $5$ minutes of computation time on commodity hardware.

We foresee that our approach could serve as a building block in more complex problem settings where we also need to optimally allocate agents to tasks (i.e., package-delivery locations) and decide the order in which to execute them. For instance, in a manner similar to our previous work~\cite{choudhury2021efficient}, we could use a bi-level approach where the upper layer allocates jobs to the agents and a lower-level planner---a coordinated MAPF solver in our case---executes the allocation in a receding-horizon fashion.
\vspace{5pt}

\noindent\textbf{Layout.} In~\Cref{sec:related} we review prior related fundamental approaches and state-of-the-art applications. In~\Cref{sec:problem} we introduce basic notation and definitions and describe our problem setting of coordinated MAPF. In~\Cref{sec:approach} we describe our two-stage approach for coordinated MAPF in detail and discuss our extensive experimental results in~\Cref{sec:results}. We conclude by summarizing our work and outlining future research directions in~\Cref{sec:conclusion}.


%% file: tex/related.tex
\section{Related Work}
\label{sec:related}

We briefly review three related areas of prior research: the general multi-agent path finding problem, peer-to-peer ridesharing algorithms, and coordinated logistics with drones and trucks.

\subsection{Multi-Agent Path Finding}
\label{sec:related-mapf}
The problem of planning paths for a team of agents subject to domain-specific inter-agent constraints (e.g., collision avoidance) is known as Multi-Agent Path Finding or MAPF~\cite{yu2013structure}. MAPF is a sub-class of the more general Multi-Agent Planning problem~\cite{torreno2017cooperative}.
Though the underlying MAPF problem is computationally hard, the research community has developed several effective search-based solvers that work well in practice~\cite{sharon2012conflict,barer2014suboptimal,felner2017search}. 

Most MAPF algorithms are evaluated on grid-worlds where agents can move step-wise along the four cardinal directions, rather than on large-scale road networks~\cite{stern2019multi}. They are also designed for avoiding collisions, not enabling agents to temporarily coordinate by actively sharing their locations. There are two relevant exceptions. First, a recent paper developed a bounded-suboptimal MAPF approach that routes drones over a time-dependent ground transit network~\cite{choudhury2021efficient}; however, the ground vehicles are fixed and not controllable. Second, an algorithmic extension to the classical MAPF formulation enables agents to explicitly cooperate~\cite{greshler2021cooperative}. But this approach assigns agents to cooperative tasks in advance and does not require them to simultaneously traverse a shared path.

\subsection{Peer-to-Peer Ridesharing}
\label{sec:related-p2p}

The recent shift towards shared-use mobility services has motivated the study of peer-to-peer ridesharing. This problem involves matching drivers to passengers with similar itineraries, such that the former can share their trips with the latter~\cite{agatz2012optimization}. A taxonomy of ridesharing variants has emerged, based on the type of matching required~\cite{tafreshian2020frontiers}; our problem in this paper can be considered a fixed-role many-to-many matching variant~\cite{masoud2017decomposition}, where the trucks are the `drivers' and drones are the `passengers'. Several ridesharing algorithms provide useful foundations for us, such as one that geographically partitions a large-scale road network~\cite{pelzer2015partition}, one that partitions an intermediate trip graph datastructure to decompose the problem~\cite{tafreshian2020trip}, and one that incorporates return restrictions on drivers in an integer linear program~\cite{chen2019ride}. However, a common challenge with all of them is that their objective functions only consider the distance traveled by the driver vehicles, and possibly wait time for the passengers, but not the distance that passengers must traverse to get to the drivers. Another challenge is that most approaches frame and solve a mathematical program, which scales poorly to complex multi-agent path finding problems~\cite{bartak2017modeling}.

\subsection{Coordinated Drone-Truck Routing}
\label{sec:related-drone-truck}

The idea of pairing drones with trucks for last-mile delivery and logistics has been partially explored. The flying-sidekick traveling-salesman problem was formulated to model a single truck-drone pair making a set of deliveries, with the drone leaving and returning to the truck at various points~\cite{murray2015flying}. A range of optimization approaches have been developed for this flying sidekick problem~\cite{AlenaETAL18}, including an extension that considers multiple drones~\cite{murray2020multiple}. Unfortunately, they do not scale well with scenario size and are only applied for a small number of trucks and drones.

Several other works address similar settings; a genetic algorithm optimizing a truck-drone pair in a tandem delivery network~\cite{ferrandez2016optimization}, a geometric approach that relies on Euclidean plane analysis~\cite{carlsson2018coordinated}, a sequential decision-making model that assumes geographical districting~\cite{ulmer2018same}, and a drone scheduling routine for given truck routes~\cite{boysen2018drone}. All of those approaches consider a small number of agents (typically one drone and one truck) and allow the drones to move freely in the plane, which can be unrealistic. Lastly, we mention that control aspects of drone landing on a moving truck have been explored recently~\cite{Haberfeld.ea.2021}.  


%% file: tex/problem.tex
\section{Problem Formulation}
\label{sec:problem}
We control a centralized fleet of agents comprising $\numtrucks$ trucks and $\numdrones$ drones. Each agent is assigned an index $i$ from the set $\A=\A^\tr\cup \A^\dr$, where $\A^\tr:=\{1,\ldots, \numtrucks\}$ and $\A^\dr:=\{\numtrucks+1, \numtrucks+\numdrones\}$ denote truck and drone indices, respectively. The agents operate on a shared road network, represented as a directed graph $\G = \left(\V, \E\right)$ with two types of edges.  An edge $(v,v')\in \E$ where $v\neq v'$ represents a traversal of a physical road segment.

Each edge $e \in \E$  has two cost values $c^\tr(e)$ and $c^\dr(e)$ representing the travel cost incurred by a truck and drone, respectively. In both cases, we set $c^\tr(e)$ and $c^\dr(e)$ to the physical distance of the corresponding road link, but the two quantities could be different. The drone incurs cost $c^\dr(e)$ while traversing $e$ only if it is flying along the edge. Our objective, discussed in detail below, is to optimize the total travel cost incurred by all agents. Edges are also annotated with travel times that depend on the type of agents using them, based on an average traversal speed. We consider a discrete time setting in this work, where $t^\tr(e)$ and $t^\dr(e)$ denote the integer traversal time for trucks and drones, respectively.

\subsection{Truck and drone paths}
Each agent $i \in \A$ is assigned start and goal nodes $s_i, g_i \in \V$. For simplicity, we assume that all the agents begin their journey at time step $0$. We could account for different start times without losing generality, by including a new zero-cost edge for each agent with traversal time equal to its start time.
We must compute a set of paths that move agents from their start to goal nodes over the graph $\G$, such that the total travel cost across all agents is minimized. 

Drones can use trucks as temporary modes of transit for one or more edges to save on travel cost. They may only board or alight at nodes in the road network graph. Each truck has a maximum carrying capacity for drones. In our experiments, all trucks are homogeneous and have the same capacity $C \geq 1$, though our approach could accommodate varying truck capacities. 

Next, we describe the solution paths for trucks, which encode their traversal over the road network graph.  
Given a truck $i\in \A^\tr$, its solution path $\pi_i=\left(e^0_i,e^1_i,\ldots,e^{\ell_i}_i\right)$ is a sequence of edges, for some $\ell_i\geq 1$, where for every $0\leq j\leq \ell_i$ it holds that $e^j_i\in \E$, and $o(e^0_i)=s_i, d(e^{\ell_i}_i)=g_i$, where $o(e)$ and $d(e)$ denote the origin and destination of a given edge $e\in \E$. Additionally, a path must satisfy connectivity constraints between the edges, i.e., $d(e_i^j)=o(e_i^{j+1})$ for every $0\leq j<\ell_i$. 
In addition to encoding the truck's location in space, the path $\pi_i$ also implicitly describes its departure and arrival time over the edges. In particular, $\tau^j(\pi_i)$ denotes the departure time of edge $e_i^j\in \pi_i$, where $0 \leq j < \ell_i$, and is defined as 
\begin{align}
\tau^j(\pi_i):=t^\tr(e_i^j)+\tau^{j-1}(\pi_i),
\end{align} 
where $\tau_0(\pi_i):=0$.    The path cost incurred by the truck is simply the sum of truck travel costs $c^\tr(\pi_i):=\sum_{0\leq j\leq \ell_i}c^\tr(e^{j}_i)$.

We define solution paths for drones similarly, except that these paths also describe whether drones use trucks for some trip segments. For drone $i\in \A^\dr$, its solution is a path  $\pi_i=\left(e^0_i,e^1_i,\ldots,e^{\ell_i}_i\right)$ for some $\ell_i\geq 1$ and an assignment sequence $A_i=\left(a_i^0,a_i^1,\ldots,a_i^{\ell_i}\right)$. The start-goal and continuity constraints imposed on $\pi_i$ are the same as those for trucks, as are the arrival and departure times of the drone path edges.

The assignment sequence $A_i$ is defined as follows. For a given solution segment $0\leq j\leq \ell_i$, the value $a_{i}^j\in \A^\tr\cup \{\bot\}$ describes whether drone $i$ is riding a truck $a^j_{i}\in \A^\tr$ or flying (in which case $a^j_{i}=\bot$), when traversing the edge $e\in \E$. Assuming that the assignment sequence $A_i$ is valid with respect to the trucks used along the route (defined below), the cost of the drone solution $(\pi_i,A_i)$ is computed as
\begin{align}
    c^\dr(\pi_i,A_i):=\sum_{0\leq j\leq \ell_i}\indicator_{a_i^j=\bot}c^\dr(e_i^j),
\end{align} 
where $\indicator_{a_i^j=\bot}$ returns $1$ if $a_i^j=\bot$, and otherwise returns $0$. That is, the value $c^\dr(\pi_i,A_i)$ \textit{sums up the cost along edges for which the drone does not ride on a truck}.

A global solution $\S$ to our problem is a collection of solutions over all agents, i.e., $\S:=\bigcup_{i\in \A^\tr}\{\pi_i\}\cup \bigcup_{i\in \A^\dr}\{(\pi_i,A_i)\}$. The solution $\S$ is \emph{valid} if it satisfies the following two conditions on the coordination between trucks and drones. First, fix a drone $i\in \A^\dr$ and fix a segment $0\leq j\leq \ell_i$ where it is assigned to some truck $i':=a_i^j\in \A^\tr$. Then the edge $e_i^{j}\in \pi_i$ along the drone's path must also be part of the truck's path $\pi_{i'}$, whose departure time must also align with that of the drone. That is, there exists a truck segment $0\leq j'\leq \ell_{i'}$ such that $e_i^{j}=e_{i'}^{j'}$ and $\tau_j(\pi_i)=\tau_{j'}(\pi_{i'})$. The second condition requires that a truck's capacity will not be exceeded. In particular, fix a truck $i\in \A^\tr$ and a solution segment $0\leq j\leq \ell_i$. Then it must hold that
\begin{align}
    \left|\left\{i' \in \A^\dr \mid  \exists 0\leq j'\leq \ell_{i'}, \tau_{j'}(\pi_{i'})= \tau_{j}(\pi_i) \wedge a_{i'}^{j'}=i\right\}\right|\leq C.
\end{align}

We are ready to state our problem.
\begin{problem}[Coordinated MAPF]\label{prb:main}
  Given a set of agents $\A=\A^\tr\cup  \A^\dr$, road graph $\G$, edge cost function $c$, edge traversal-time function $t$, we wish to find a valid global solution $\S:=\bigcup_{i\in \A^\tr}\{\pi_i\}\cup \bigcup_{i\in \A^\dr}\{(\pi_i,A_i)\}$ minimizing the total cost
  \begin{align}
      c(\S):=\sum_{i\in \A^\tr}c^\tr(\pi_i) +  \sum_{i\in \A^\dr}c^\dr(\pi_i,A_i).
  \end{align}
\end{problem}

\subsection{Discussion}

We discuss some computational aspects of our problem.
Integer programming would scale poorly, even for only one truck and one drone~\cite{murray2015flying}. Coordinating multiple drones and trucks increases the nominal decision space by orders of magnitude through two axes:
(i) all possible matchings of drones to trucks based on capacity (i.e., all ways to distribute $n$ drones across $m$ trucks where each truck can have up to $C$ drones); (ii) for each drone-truck pairing, all possible intermediate start and end points of the truck route, and the various routes a truck can take.
We aslo need to account for conflicts arising from violating vehicle capacity constraints. Those observations suggest that our problem is more difficult than the classical MAPF problem, which is NP-hard~\cite{yu2013structure}. But we defer the study of the complexity of Coordinated MAPF for future work.

We now discuss our modeling assumptions. First, we use a shared road graph for trucks and drones. For urban areas with high-rises and no-fly zones, restricting drones to fly over the road network rather than point-to-point between any two locations is a reasonable design choice.
Second, of the two popular MAPF objective functions, \textit{sum-of-costs} and \textit{makespan}, we choose to minimize the former rather than the latter (the makespan of a MAPF solution is the maximum path cost for any agent). 

In our setting, \textit{sum-of-costs better reflects the gains from having drones use trucks as transit}; e.g., if the maximum-cost path in a solution was that of some truck whose start and goal were disproportionately far apart, then the makespan of the solution would only depend on the cost of that worst path (unlike the sum-of-costs metric); no further optimizing of the other drones and trucks to save on drone flight cost would be incentivized, even though that would have led to real-world benefits. In any case, we considered makespan in a related earlier work where it was more appropriate~\cite{choudhury2021efficient}.
Finally, we only include travel distance in our objective and not the elapsed travel time to avoid arbitrary scaling between the two physical quantities. Vehicle distance traveled is a standard objective for the ridesharing problem~\cite{tafreshian2020frontiers}.


%% file: tex/approach.tex
\section{Coordinated Drone-Truck MAPF}
\label{sec:approach}

Most effective multi-agent planning approaches rely on the system being loosely coupled~\cite{brafman2008one}. In the context of MAPF (a sub-class of multi-agent planning), \textit{a loosely coupled multi-agent system} is one where the optimal path for an individual agent mostly does not interact or coordinate with that of other agents, and when it does, the extent of interaction required is small compared to the overall path length. For example, Conflict-Based Search or CBS~\cite{sharon2012conflict} is an influential MAPF algorithm that works by computing individual paths independently for each agent with a search-based method like A*~\cite{hart1968formal} and resolving any conflicts or inter-path constraint violations in a structured hierarchical manner. However, it only works well on problems where there are not many conflicts, i.e., the underlying multi-agent system is loosely coupled~\cite{Gordon.ea.21}.

\begin{figure*}[t]
    \centering
    \begin{subfigure}{0.45\textwidth}
        \centering
        \includegraphics[width=0.95\textwidth]{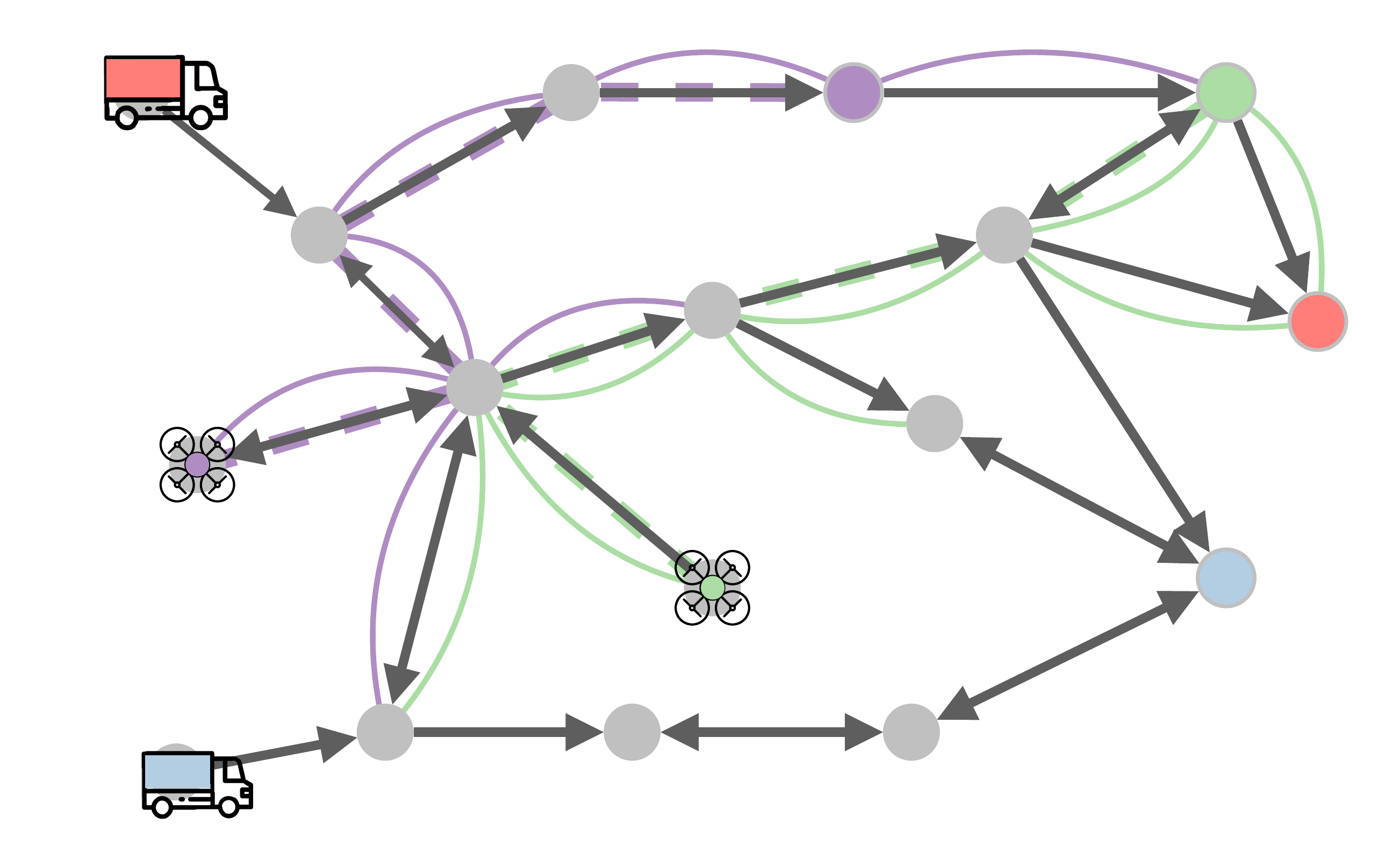}
        \label{fig:stage1-mapf}
    \end{subfigure}
    \begin{subfigure}{0.45\textwidth}
        \centering
        \includegraphics[width=0.95\textwidth]{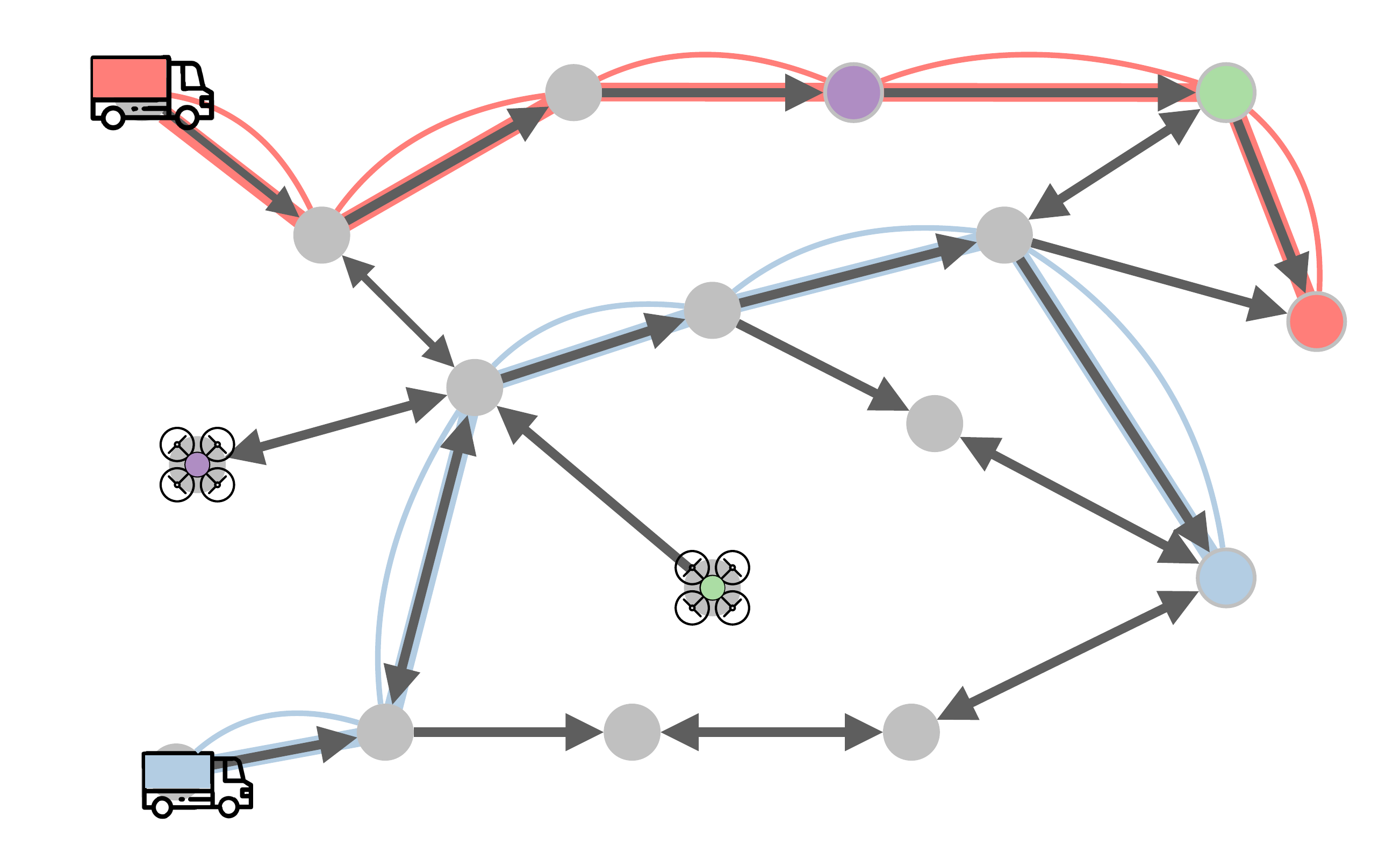}
        \label{fig:stage2-mapf}
    \end{subfigure}
    \caption{The two MAPF sub-problems that we need to solve. (Left) In Stage 1, we take initial drone shortest paths (dashed purple and green) and add drone-annotated weight-discounted copies of nearby road edges (coloured thinner edges; for clarity, we only show a subset of all possible such copies). We then solve a MAPF problem for trucks on this augmented graph. (Right) In Stage 2, we take the truck paths computed in Stage 1 (solid red and blue) and create zero-weight transit edge copies from them. We then solve a MAPF problem for drones on this road-and-transit graph.
    }
    \label{fig:stage-details}
\end{figure*}

In our problem, the ability of drones to coordinate with trucks when convenient \textit{vastly increases the amount of coupling in the system} and thereby the complexity of the MAPF problem. If drones incur no cost when using trucks as transit, then their optimal paths are highly dependent on the set of truck paths. On the other hand, the set of individual shortest paths for trucks may not yield a useful transit network for the drones. We noted this increased complexity in the previous section, through the orders-of-magnitude larger decision space. But the tighter coupling affects even search-based solvers like CBS, which are less sensitive to dimensionality than integer programming methods.

The tight coupling and large-scale road network in our setting make it intractable to solve optimally or bounded-suboptimally. Therefore, \textit{we decouple the overall MAPF problem into two distinct MAPF sub-problems} and solve them in stages (\Cref{fig:approach}). In Stage 1, we compute a set of truck paths $\{\pi_i^1\}_{i\in \A^\tr}$ that drones are likely to use as transit. We do so by generating a MAPF instance for the trucks where their travel cost is discounted when they get close to \emph{nominal} drone routes, i.e., the shortest drone paths over the road network.
In Stage 2, we design another MAPF instance to route the drones. We augment the road graph $\G$ to keep track of the motion of the trucks along $\{\pi_i^1\}_{i\in \A^\tr}$ and allow drones to ride on the trucks for some segments of their trips and fly for the others. 
After both stages, we re-route any trucks that were not used by a single drone to their original shortest path on the road network, if their Stage 1 paths deviated from their shortest path.

\subsection{Stage 1: MAPF for Trucks}
Our problem imposes no system constraints between any two truck paths. If we ignore drones (whose paths will be computed in Stage 2), the solution to the MAPF problem for the trucks alone is trivial: the set of shortest road network paths for each individual truck. However, this set of truck paths may be ill-suited for the drones to use as transit and may yield a poor-quality downstream solution. Therefore, \textit{we seek a set of truck paths that drones are likely to benefit from in Stage 2}. To inform this search, we first consider a set of one or more start-goal flight paths for each drone over the road network. In our case, we use the shortest flight path $\pi_i^0$ for each drone $i\in \A^\dr$ along $\G$, but this set could capture other properties like geographical coverage or path diversity~\cite{bast2013result}.

Next, we describe an instance of the MAPF problem whose solution would allocate trucks to routes that could potentially benefit the drones. We describe the graph $\G^\tr=(\V^\tr,\E^\tr)$ used in this stage. The graph $\G^\tr$ augments the road network graph $\G$ to encourage trucks to take paths that are close to the initial drone flight paths. In particular, the vertices $\V^\tr$ are equal to $\V$. The edge set $\E^\tr$ generalizes the edge set $\E$, i.e., $\E\subset \E^\tr$. For each drone $i\in \A^\dr$, we consider all edges $e \in \E$  within $K\geq 1$ hops (or edge traversals) of its flight path $ \pi_i^0$. For each such edge $e \in \E$, we \textit{create a copy $e^i_k \in \E^\tr$ annotated with the drone and the hop $0\leq k\leq K$}; the same road network in $\E$  can have multiple copies in $\E^\tr$ if it is close to multiple nominal drone paths. See Figure~\ref{fig:stage-details} (Left) for illustration. 

The weights of the road edges $e\in \E$ reflect their truck cost, i.e., $w^\tr(e):=c^\tr(e)$. For each copy $e^i_k$ we discount the original edge truck cost $c^\tr(e)$ by $\gamma(k) \in [\frac{1}{2}, 1)$ depending on the hop distance $k$ between the edge and the corresponding drone flight path, i.e., $w^\tr(e^i_k):=c^\tr(e) \cdot \gamma(k)$. We want $\gamma(0) = \frac{1}{2}$ to reflect that the drone need not deviate from its shortest path at all if the truck takes that edge, thus halving the effective edge cost.
The choice of discounting function is a heuristic. For our experiments in~\Cref{sec:results}, we use $\gamma(k) = \frac{1}{2} \cdot (1 + \mathrm{tanh}(k))$. We also run an ablation study with the \textit{sigmoid} function, i.e. $\gamma(k) = \frac{1}{1 + e^{-k}}$ in the extended version of our paper~\cite{choudhury2021coordinated}. Both satisfy our desiderata that $\gamma(0) = \frac{1}{2}$ and that $\gamma(k)$ approaches $1$ as $k$ increases. The maximum hop distance ($K$) the drone flight path is $3$ hops in all cases; we ran some offline ablations, found negligible change for greater values of $K$, and omit those results in the interest of space.

The road network graph, augmented by discounted weight copies, is the underlying pathfinding graph $\G^\tr$ for the MAPF problem with trucks (\Cref{fig:stage-details}; left). The shortest paths for trucks (with respect to the weights $w^\tr$) on this augmented graph $\G^\tr$ may deviate from their original shortest paths (with respect to weights $c^\tr$) on the graph $\G$, in favor of edges with drone-annotated copies and thus lower weights. The \textit{truck paths on the augmented road graph do have pairwise constraints}, unlike those on the base graph. To reduce the chance that several trucks are assigned to ``assist'' the same drone, we impose a capacity constraint of one, i.e., $\textup{cap}(e^i_k)=1$ on every edge $e^i_k\in \E^\tr$ (but do not restrict capacities of edges $e\in \E^\tr$ not associated with drones). In the terminology of classical MAPF, two trucks are in \emph{conflict} if they use the same drone-annotated weight-discounted edge copy $e^i_k$. Note that we do not forbid multiple trucks being assigned to copies $e^i_k,e^i_{k'}\in \E^\tr$ annotated with the same drone $i$ but different hops $k$ and $k'$. This choice keeps the Stage 1 runtime low but may introduce some known inefficiency, as the drone $i$ can only use one truck at a time. 


Given graph $\G^\tr$, weights $w^\tr$, truck capacities and start-goal nodes $\{s_i,g_i\}_{i\in \A^\tr}\subset \V^\tr$, we solve a MAPF problem with the objective of minimizing the total weight $w^\tr$ incurred by all trucks. The outcome of Stage 1 is a set of truck paths $\{\pi_i^{1}\}_{i\in \A^\tr}$. Capacity conflicts between truck paths do not reflect any physical constraints, in contrast to most MAPF applications. Rather, they encode our design choice that truck paths be useful for drones downstream. Again, since a particular drone can only use one truck at a time on transit, \textit{if two trucks deviate to use the same drone-annotated edge copy, one of those deviations is likely to be wasted in Stage 2}. 

\subsection{Stage 2: MAPF for Drones}
\label{sec:approach-mapf-drones}

Stage 1 yields a set of truck paths $\{\pi_i^{1}\}_{i\in \A^\tr}$ over the road network $\G$, because a truck using an edge copy $e^i_k\in \E^\tr$ is essentially using $e\in \E$. Whether a truck path used drone-annotated discounted-weight edge copies is now irrelevant as the discounting was simply a heuristic to guide trucks closer to the nominal drone flight paths. In Stage 2, we will fix the truck paths, keep track of their locations in time and space, and solve a MAPF problem for drones that can use trucks as transit, subject to capacity constraints. This stage builds upon recent  work for routing drones over ground transit~\cite{choudhury2021efficient}, but differs in how it requires drones to operate on the road network rather than fly point-to-point between locations.

For the Stage 2 MAPF problem, we augment the original road network with the transit network derived from truck paths (\Cref{fig:stage-details}; right panel) to yield the graph $\G^\dr=(\V^\dr,\E^\dr)$. Each truck path is a sequence of road network edges. For each road edge of a given truck path, we add a copy to $\G^\dr$ that we call a \textit{transit edge}. Every transit edge $e\in \E^\tr$ is annotated with the corresponding truck, has drone capacity $C$, and has zero weight $w^\dr(e)=0$ because the drone incurs no distance cost when using transit. We set the weight of non-transit edges $e\in \E$ to be $w^\dr(e)=c^\dr(e)$. Since we omit elapsed time in the objective and assume that drones and trucks can slow down as needed to wait for the coordinating agent, any drone-truck connection can be made in principle. 

In contrast to Stage 1, the conflicts here do represent physical constraints, i.e., the maximum drone-carrying capacity of the truck. Drone paths may \emph{conflict} with each other if more than $C$ drones use the same transit edge. Given the graph $\G^\dr$, weights $w^\dr$, and the above constraints, we have a well-defined MAPF problem for Stage 2, where we seek to compute a set of drone paths, each with a combination of road and transit edges (although there may be drone paths that do not take transit at all). Our objective is to minimize the sum of drone path costs, given that only road edges incur cost due to the distance of the corresponding link.

\subsection{Solving stage-wise MAPF problems}
\label{sec:approach-solve}

\begin{algorithm}[t]
\caption{Pseudocode of two MAPF techniques}
\begin{algorithmic}[1]
    \Statex \textbf{Input:} MAPF Graph $G$, $\ell$ agents
    \Statex
    \Procedure{ConflictBasedSearch}{}
        \State Initialize constraint tree node $A$ with $A.constr = \varnothing$
        \For{$i = 1 \text{ to } \ell$} \Comment{Any ordering}
            \State $A.soln[i] \gets \mathrm{ShortestPath}(G,i,A.constr)$ \Comment{A*}
        \EndFor
        \State $A.cost \gets \mathrm{SumOfCosts}(A.soln)$
        \State Insert $A$ into $\textsc{Open}$  \Comment{Higher-level open list}
        \While{\textsc{Open} is not empty}
            \State $B \gets \mathrm{PopBest}(\textsc{Open})$    \Comment{Min. cost candidate}
            \If{$B.soln$ has no conflicts}
                \State \textbf{return} $B.soln$ \Comment{Best valid solution}
            \EndIf
            \For{all conflicts $(i,c) \in B.soln$}
                \State $C \gets B \cup \mathrm{UpdateConstraints}(B.constr, c)$
                \State $C.soln[i] \gets \mathrm{Path}(G,i,C.constr)$
                \State Insert $C$ into \textsc{Open}
            \EndFor
        \EndWhile
    \EndProcedure
    \Statex
    \Procedure{PrioritizedPlanning}{}
        \State Initialize $S$ with $S.soln =\varnothing$ and $S.constr = \varnothing$
        \For{$i = 1 \text{ to } \ell$} \Comment{Priority ordering}
            \State $S.soln[i] \gets \mathrm{ShortestPath}(G,i,S.constr)$
            \State $\mathrm{UpdateConstraints}(S.constr, S.soln[i])$
        \EndFor
        \State \textbf{return} $S.soln$
    \EndProcedure
\end{algorithmic}
\label{alg:MAPF-solvers}
\end{algorithm}

The prior work on drone-transit routing that we build upon~\cite{choudhury2021efficient} used Enhanced CBS or ECBS~\cite{barer2014suboptimal} for the MAPF problem. CBS is a hierarchical algorithm, where the multi-agent level defines inter-path and per-path constraints on the single-agent level. The single-agent level computes optimal paths that satisfy the respective per-path constraints. If two or more single-agent paths conflict with each other, i.e., violate any shared constraints, the multi-agent level imposes more constraints to resolve the conflict, and reruns the single-agent level for the conflicting agents.

ECBS uses bounded-suboptimal Focal Search~\cite{pearl1982studies} instead of optimal A* at both levels. It can be orders of magnitude more efficient than CBS, especially on MAPF problems with many more conflicts, i.e., that reflect a more tightly coupled multi-agent system. We also implement and use ECBS for our MAPF problems in both stages. However, for larger numbers of cars and drones ECBS would have far too many conflicts to resolve and timeout before returning a solution (as we shall see in~\Cref{sec:results}). Moreover, conflict resolution is particularly expensive in our setting as the truck capacities are greater than one, and resolving them generates a large number of constraints. A conflict generates constraints for all subsets of excess drones, i.e., if a transit edge has capacity $C$, and $C'>C$ drones choose to use it, then $C'$-choose-$C$ constraints are generated in the higher-level search tree, where all $C$-subsets of the $C'$  drones are restricted from using that transit edge. A similar setting with capacities is discussed in our earlier work~\cite{choudhury2021efficient}.
A version of CBS tailored for MAPF with capacity constraints was recently developed~\cite{surynek2019multi}. But developing vanilla MAPF solvers is not the focus of our work and we defer its implementation for future research.

The challenges of using ECBS for larger problem settings motivates us to also consider Prioritized Planning (PP)~\cite{silver2005cooperative}. Here, we plan paths for agents one-by-one (using A*) based on some priority ordering. After planning each agent's path, PP analyzes the edges along it and updates any MAPF constraints for subsequent paths. In both our stages, these update rules depend on the conflict criteria between paths. In Stage 1, if a truck path uses any drone-annotated edge copies, then those copies are removed and unusable for subsequent trucks. In Stage 2, each time a drone uses a specific transit edge, its capacity is reduced by one; when a transit edge reaches zero capacity, it can no longer be used by subsequent drones.  Once a drone-annotated or transit edge is used up in the respective stages, later agents are not allowed to use them. Note that PP circumvents the conflicts that CBS encounters when $C'>C$ through the imposed priority ordering.

PP has no solution quality guarantees, unlike ECBS. But it does not need to resolve conflicts and can be more efficient than ECBS in practice. \Cref{sec:results} will show how \textit{PP can solve problems intractable for ECBS and be competitive on the tractable problems}. The choice of priority ordering can impact the solution quality for PP. For Stage 2, we use the sensible heuristic of prioritizing drones whose shortest paths on the road have higher cost, as they are most likely to benefit from using transit. No such-clearly motivated heuristic exists for the prioritizing the order of trucks paths in Stage 1, so we impose an arbitrary ordering based on truck IDs.

\Cref{alg:MAPF-solvers} contains high-level pseudocode for CBS. It maintains a higher-level constraint tree whose root node $A$ is initialized with an empty constraint set and the independent shortest paths for each agent. When the lowest-cost constraint tree node $B$ is expanded, CBS evaluates its set of paths for any conflicts and generates a child node $C$ that recomputes the path for every conflicting agent $i$ and corresponding constraint $c$. CBS continues until it yields the first conflict-free solution, which is guaranteed to be optimal. The basic structure of ECBS is the same as CBS, with a few extensions to enable more efficient behavior while sacrificing optimality for bounded-suboptimality. We omit those extensions for readability. We also sketch out the pseudocode of Prioritized Planning in~\Cref{alg:MAPF-solvers} to highlight its major differences from CBS. PP computes the shortest path of each of the $\ell$ agents one-by-one, following a given priority ordering. After obtaining the path for any agent $i$, it updates the shared set of constraints based on that path.

%% file: tex/experiments.tex
\section{Experiments and Results}
\label{sec:results}

\begin{table*}[t]
    \centering
    \caption{
    We compare the solution quality (vehicle distance traveled) and computational efficiency (plan time) of our framework against the Direct baseline. All quantities are averaged over $20$ trials. Note that by vehicle travel distance we mean the sum of truck distances and drone flight distances.
    In all cases, the standard error of the mean was less than $5\%$ of the mean, and has been omitted. \textcolor{red}{Red indicates that some of the 20 trials timed out, so the comparison is not precisely equivalent.} For the larger settings, ECBS times out on the majority of the trials, and is thus omitted.}
    \begin{tabular}{@{} llcrrrcrrccrrrcrr  @{}}
        \toprule
        &&& \multicolumn{6}{c}{Vehicle Distance Traveled (\SI{}{\kilo\metre})} &&& \multicolumn{6}{c}{Plan Time (\SI{}{\second})}\\
        \cmidrule{4-9} \cmidrule{12-17} &&&& \multicolumn{2}{c}{Cap = 5}  && \multicolumn{2}{c}{Cap = 10} &&&& \multicolumn{2}{c}{Cap = 5} && \multicolumn{2}{c}{Cap = 10} \\
        \cmidrule{5-6} \cmidrule{8-9} \cmidrule{13-14} \cmidrule{16-17} Trucks & Drones && Direct & ECBS & PP && ECBS & PP &&& Direct & ECBS & PP && ECBS & PP \\
        \midrule
        $5$ & $10$  && $113$ & $74.9$   & $80.8$ && $74.8$  & $80.8$ &&& $0.012$ & $4.76$ & $4.80$ && $4.93$ & $4.80$ \\
        $5$ & $15$  && $148$ & $93.4$   & $99.3$  && $89.8$  & $95.5$ &&& $0.013$ & $8.01$ & $6.96$ && $7.39$ & $7.03$ \\
        $5$ & $20$  && $185$ & $116$    & $127$  && $107$   & $116$  &&& $0.015$ & $10.9$   & $9.21$ && $9.79$ & $9.70$ \\
        $10$ & $20$ && $223$ & \textcolor{red}{$124$}   & $135$  && $116$  & $123$  &&& $0.021$ & \textcolor{red}{$41.6$}   & $21.6$ && $28.2$ & $22.9$ \\
        $10$ & $30$ && $295$ & --       & $187$  && $139$   & $153$  &&& $0.026$ & --   & $32.4$ && $59.8$ & $36.8$ \\
        $10$ & $40$ && $368$ & --       & $254$  && \textcolor{red}{$167$}  & $191$  &&& $0.029$ & --   & $42.7$ && \textcolor{red}{$157$}   & $48.6$ \\
        $15$ & $30$ && $333$ & --       & $197$  && \textcolor{red}{$158$}   & $167$  &&& $0.029$ & --   & $57.1$ && \textcolor{red}{$144$}   & $67.1$ \\
        $15$ & $45$ && $434$ & --       & $275$  && --      & $207$  &&& $0.039$ & --   & $78.5$ && --   & $95.7$ \\
        $15$ & $60$ && $543$ & --       & $374$  && --      & $258$  &&& $0.048$ & --   & $105$  && --   & $135$  \\
        $20$ & $40$ && $436$ & --       & $257$  && --      & $213$  &&& $0.035$ & --   & $112$  && --   & $138$  \\
        $20$ & $60$ && $576$ & --       & $367$  && --      & $265$  &&& $0.047$ & --   & $158$  && --   & $206$  \\
        $20$ & $80$ && $714$ & --       & $525$  && --      & $347$  &&& $0.058$ & --   & $210$  && --   & $296$  \\
        \bottomrule
    \end{tabular}
    \label{table:all-results}
\end{table*}

We implemented\footnote{The code is at \url{https://github.com/Shushman/AerialGroundPathFinding.jl}}  our approach and ran all simulations using the Julia programming language~\cite{Julia-2017} on a machine with \SI{128}{\gibi\byte} RAM and a \SI{2.6}{\gibi\hertz} Intel Xeon CPU.
On various problem settings, we evaluated the quality of solutions as per our optimization objective, i.e., the sum of total path  costs over all agents. We also considered the efficiency of our approach by measuring the plan time and observing how it scales with more trucks and drones. For the road graph, we used the street network of Manhattan in New York City\footnote{\url{https://www.kaggle.com/crailtap/street-network-of-new-york-in-graphml}.}, covering an area of nearly \SI{200}{\kilo\metre\squared}. This directed graph has 4426 nodes and 9604 edges; the nodes are annotated with geographical locations and the edges are annotated with the distance of the road link in kilometres. For simplicity, we use total vehicle distance traveled as the path cost metric (i.e., the sum of trucktravel  distance and drone flight distance), and we defer more sophisticated cost metrics to future work. Our underlying graph is significantly larger than in most MAPF applications~\cite{stern2019multi}.


\subsection{Solution Quality and Efficiency}
\label{sec:results-eval}

We evaluated two versions of our approach on solution quality and computational efficiency: one with Enhanced Conflict-Based Search (ECBS) for both stages and the other with Prioritized Planning (PP) instead. Since the full problem is intractable to solve jointly, we do not baseline against a Mixed Integer Linear Program approach. As a reference point, we compared against an approach that simply assigns to each truck and drone its shortest path on the road network, with no coordination among them. This baseline (that we call Direct) is much faster than our approach but has much poorer solution quality, especially with more trucks that drones can use as transit to reduce path cost.

\Cref{table:all-results} displays the results of all simulations. We varied the number of trucks and drones and considered two different drone-carrying capacities for trucks, $5$ and $10$. For each setting, we had $20$ different trials, each with different start and goal nodes for each agent. We ran all approaches on the MAPF problems defined by that trial, computed the total solution cost (vehicle distance) in kilometres and the planning time in seconds, and averaged over all trials. The standard error of the mean was less than $5\%$ of the mean in all cases, so we omitted it in the interest of space.

As expected, both ECBS and PP compute much better quality solutions than Direct, and \textit{the quality gap increases with more trucks and drones and with higher truck capacity}. Also, Direct has much lower planning time than both PP and ECBS. We mentioned earlier in~\Cref{sec:approach-mapf-drones} that ECBS has many expensive conflicts to resolve for bounded-suboptimality. PP does not explicitly need to resolve conflicts but still needs to plan over much larger graphs in both stages than the road network that Direct plans on: Stage 1 adds the drone-annotated weight-discounted edge copies and Stage 2 adds the transit network from truck paths. Even the worst plan time for PP, nearly 5 minutes, is good enough in practice for an operation horizon on the order of hours. For that setting,  PP has an absolute savings of more than \SI{350}{\kilo\metre} and a relative savings of more than $50\%$ compared to Direct.

Comparing PP to ECBS yields many important insights. The plan times start out as comparable, but PP scales much more better than ECBS with increasing trucks and/or drones, as we foreshadowed in~\Cref{sec:approach-solve}. Beyond a certain problem size, the majority of ECBS trials have too many conflicts (our threshold is $500$) and timeouts (more than 10 minutes of planning time); those entries in the table remain blank. Before those thresholds, there are settings where a minority of ECBS trials time out, making the table entry not directly comparable to PP as it averages over a subset of trials (we have marked those in red). For example, \textit{notice how the plan time for ECBS for $10$ trucks and $40$ drones shoots up to 157 seconds, i.e., nearly 3 minutes}. Problems with truck capacity 5 take longer to solve than those with capacity 10 because they are more resource constrained and hence yield more conflicts to resolve in ECBS and more constraints to update in PP.

In contrast to ECBS, PP does not have to resolve conflicts, has lower plan time for most settings (with a slight exception for 5 trucks, 10 drones, and capacity 5), and is tractable even for our largest problems. We expected it to be more efficient than ECBS, which is why we implemented it in the first place. \textit{Even more promising is the modest gap in solution quality between PP and ECBS for the settings where the latter is tractable}. The total vehicle distance traveled of the PP solution is typically between $10$ and $20\%$ higher than that of the ECBS solution (the gap for $10$ trucks and $40$ drones is higher though not entirely representative because ECBS times out on some of the $20$ problems). A possible reason for the small gap in solution quality between the two approaches is that our setting has a relatively small number of agents compared to the pathfinding graph size, which increases the space of non-conflicting agent paths and provides more flexibility. In problems that require tight coordination between agents such as automated warehouse scenarios~\cite{wurman2008coordinating}, PP may struggle to find a solution.

For both ECBS and PP, Stage 2 accounts for almost all the total plan time (over $99\%$). We expect this disparity between the stage plan times for three reasons: the number of drones to plan for is two to four times the number of trucks, constructing the composite road-and-transit network for stage 2 is itself expensive, and the second stage has more constraints and conflicts than the first.

\subsection{Ablation Study on Effect of Stage 1}
\label{sec:results-ablation-stage1}

To disentangle the relative effects of the two stages of our approach on the solution quality (vehicle distance traveled), we ran an ablation study with a modified approach. Here, the first stage does not account for the drones at all and just sets the truck routes to their shortest paths on the road network. The second stage is the same as our original approach, and plans for the drones over the composite network of the road and the truck paths as transit. We expect this modified approach to compute solutions that are better than the Direct baseline. But the nature of these performance gaps will help us understand how much of the advantage from~\Cref{sec:results-eval} is due to either stage and what future work should focus on improving.

\begin{table}[t]
    \centering
    \caption{The ablation with the modified approach Direct-PP (D-PP for short) helps disentangle the relative effects of the two stages on Vehicle Distance Traveled. The values for the PP and Direct columns are copied over from the corresponding Distance columns in~\Cref{table:all-results}.}
    \begin{tabular}{@{} llrrrcrr  @{}}
         \toprule
         &&& \multicolumn{2}{c}{Cap = 5} && \multicolumn{2}{c}{Cap =10} \\
         \cmidrule{4-5} \cmidrule{7-8} Trucks & Drones & Direct & D-PP & PP && D-PP & PP\\
         \midrule
         $5$  & $20$    & $185$ & $130$     & $127$   && $119$  & $116$ \\
         $10$ & $40$    & $368$ & $257$     & $254$   && $200$  & $191$ \\
         $15$ & $60$    & $543$ & $376$     & $374$   && $273$  & $258$  \\
         $20$ & $80$    & $714$ & $526$     & $525$   && $354$  & $347$   \\
         \bottomrule 
    \end{tabular}
    \label{table:stage1_ablation}
\end{table}

For the ablation study, we considered problem settings with a higher ratio of the number of drones to the number of trucks ($4$). In~\Cref{sec:results-eval}, we had considered smaller ratios as well ($2$ and $3$) to highlight the trends in solution quality and plan time with an increasing drone/truck ratio, but in practice we would prefer higher ratios as trucks are more expensive to operate than drones and contribute to ground congestion. The modified approach uses direct truck shortest paths for Stage 1 and PP for Stage 2; we call it Direct-PP (D-PP in short). We compare its solution quality against that of PP for both stages over the problem settings in~\Cref{table:stage1_ablation}, averaged over $20$ trials. The entries in the PP columns are copied over from the corresponding columns in~\Cref{table:all-results}.

The performance gap between D-PP and PP appears to be quite modest for most settings, particularly with capacity $5$. Behind the low average differences, however, are \textit{specific instances where PP yields a significantly better solution than D-PP}, e.g., in a few of the settings with $15$ drones, $60$ trucks, and capacity $10$, PP's solution was more than $20\%$ better than that of D-PP; we found a similar gap for a few of the settings with $10$ drones, $40$ trucks, and capacity $10$.
Of course, PP can be arbitrarily sub-optimal depending on the priority ordering (and we choose an arbitrary ordering for Stage 1). Future work could investigate various modifications and improvements to Stage 1: a more sophisticated priority ordering, a different set of nominal drone flight paths to use as the basis for weight-discounted copies of nearby road edges, or even a hybrid of ECBS for Stage 1 (which takes less time) and PP for Stage 2, to yield potentially better overall solution quality without an unacceptably large decrease in computational efficiency.

%% file: tex/conclusion.tex
\section{Conclusion}
\label{sec:conclusion}

We introduced the problem of coordinated routing for drones and trucks over a common large-scale road network, where the former can use the latter as modes of transit to reduce total vehicle distance traveled. We explained how this problem is significantly more complex than prior work on MAPF. Our comprehensive algorithmic framework elegantly decouples the intractable overall problem into stage-wise multi-agent path finding sub-problems that it solves for trucks and drones respectively. In practice, it yields significant distance savings compared to independently operating trucks and drones (more than $50\%$), within a reasonable computation time (up to $5$ minutes) on the large-scale Manhattan road network.

Several interesting operational extensions emerge for future work, including equipping trucks with charging docks for the drones and allowing drones to drop packages onto moving trucks. In addition, we could further enhance the performance of our framework through an iterative process that improves truck and drone routes repeatedly, replans online in receding-horizon fashion, and considers hybrid methods using different heuristics in each stage. Finally, for practical applications, it would be useful to extend our method to the lifelong MAPF setting~\cite{ma2017lifelong} where agents receive new tasks when they complete their current one, and to consider more complex trajectory-level issues of the drone routes, such as kinematic constraints.

